\documentclass[10pt,twocolumn,letterpaper]{article}

\usepackage{cvpr}
\usepackage{times}
\usepackage{epsfig}
\usepackage{graphicx}
\usepackage{amsmath}
\usepackage{amssymb}
\usepackage{booktabs}
\usepackage{diagbox}
\usepackage{makecell,multirow,diagbox}

\usepackage[breaklinks=true,bookmarks=false]{hyperref}

\cvprfinalcopy 

\def\cvprPaperID{1469} 

\begin{document}

\title{Detecting Curve Text in the Wild: New Dataset and New Solution}

\author{Yuliang Liu, Lianwen Jin$^{*}$, Shuaitao Zhang, Sheng Zhang\\
College of Electronic Information Engineering\\
South China University of Technology \\
{\tt\small liu.yuliang@mail.scut.edu.cn; $^{*}$lianwen.jin@gmail.com}
}

\maketitle

\begin{abstract}
  Scene text detection has been made great progress in recent years. The detection manners are evolving from axis-aligned rectangle to rotated rectangle and further to quadrangle. However, current datasets contain very little curve text, which can be widely observed in scene images such as signboard, product name and so on. To raise the concerns of reading curve text in the wild, in this paper, we construct a curve text dataset named CTW1500, which includes over 10k text annotations in 1,500 images (1000 for training and 500 for testing). Based on this dataset, we pioneering propose a polygon based curve text detector (CTD) which can directly detect curve text without empirical combination. Moreover, by seamlessly integrating the recurrent transverse and longitudinal offset connection (TLOC), the proposed method can be end-to-end trainable to learn the inherent connection among the position offsets. This allows the CTD to explore context information instead of predicting points independently, resulting in more smooth and accurate detection. We also propose two simple but effective post-processing methods named non-polygon suppress (NPS) and polygonal non-maximum suppression (PNMS) to further improve the detection accuracy. Furthermore, the proposed approach in this paper is designed in an universal manner, which can also be trained with rectangular or quadrilateral bounding boxes without extra efforts. Experimental results on CTW-1500 demonstrate our method with only a light backbone can outperform state-of-the-art methods with a large margin. By evaluating only in the curve or non-curve subset, the CTD + TLOC can still achieve the best results. Code is available at {\color{blue} https://github.com/Yuliang-Liu/Curve-Text-Detector}.
\end{abstract}

\section{Introduction}

Text in the wild conveys valuable information, which can be used for real-time multi-lingual translation, behavior analysis, product identification, automotive assistance, etc.
\begin{figure}[htb]
\begin{minipage}[c]{01\linewidth}
  \centering
  \centerline{\includegraphics[width = 8.0cm, height = 3.5cm]{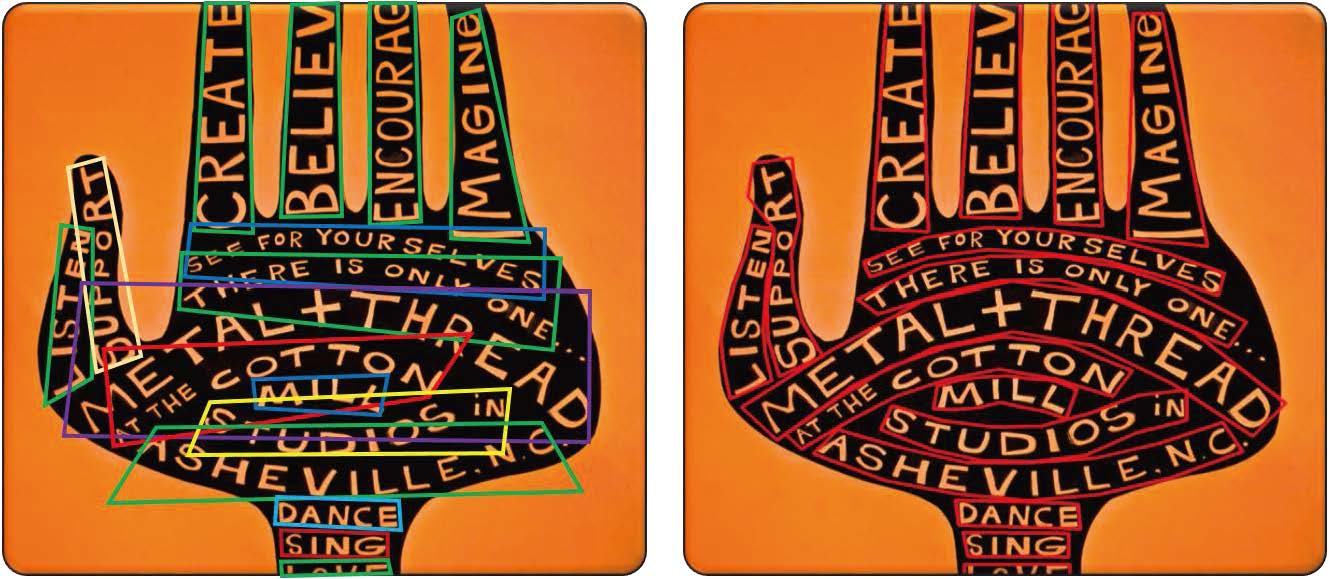}}
  \centerline{\small{(a) Curve bounding box can avoid needless overlap. }}\medskip
\end{minipage}
\vfill  
\begin{minipage}[c]{01\linewidth}
  \centering
  \centerline{\includegraphics[width = 8.0cm, height = 3.5cm]{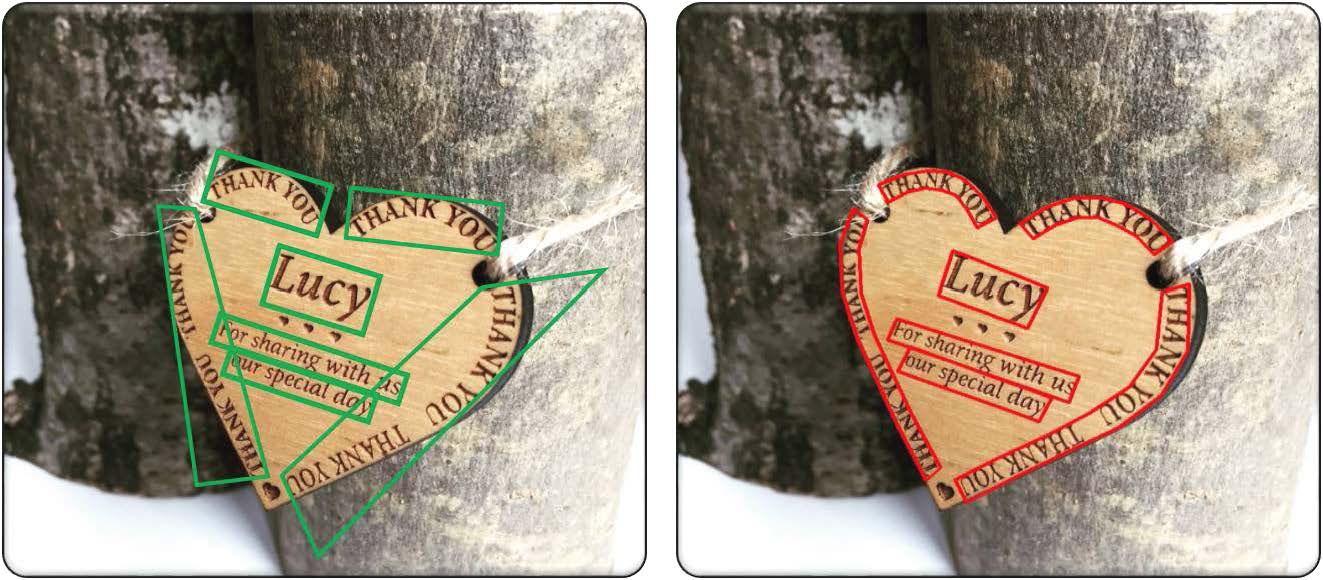}}
  \centerline{\small{(b) Curve bounding box can reduce redundant background noise.}}\medskip
\end{minipage}
\vfill  
\begin{minipage}[c]{01\linewidth}
  \centering
  \centerline{\includegraphics[width = 8.0cm, height = 3.5cm]{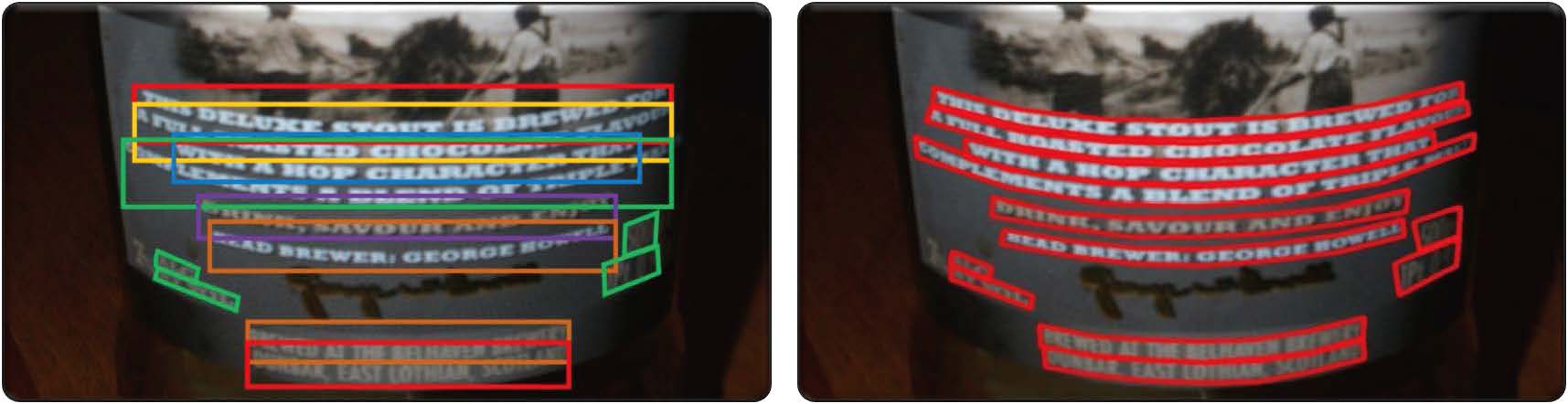}}
  \centerline{\small{(c) Curve bounding box is easier to recognize text. }}\medskip
\end{minipage}

\caption{Comparison of quadrilateral bounding box and rectangular bounding box for localizing texts. Left: using quadrilateral label. Right: using polygonal label.}\label{fig:background}
\end{figure}
Recently, the emergence of many text datasets, which are constructed on specific tasks and scene, have contributed to great progresses of text detection and recognition methods. Interestingly, it is observed that labels of text bounding boxes from emerging datasets also develop from rectangle to flexible quadrangle. For example, horizontal rectangular labels in ICDAR 2013 ``Focus Scene Text''~\cite{Karatzas2013ICDAR} and SVT~\cite{wang2010word}; rotated rectangular labels in MSRA-TD500~\cite{Yao2012Detecting} and USTB-SV1K~\cite{Yin2015Multi}; four points labels in ICDAR 2015 ``Incidental Text''~\cite{karatzas2015icdar}, RCTW-17~\cite{Shi2017ICDAR2017} and recent MLT competition dataset~\cite{mlt2017icdar}. Similarly, the advancements of scene text detection methods also change from axis-aligned rectangle based to rotated rectangle based and to quadrangle based. It was indicated in~\cite{liu2017deep} that once the bounding box becomes tighter and flexible, it can improve the detecting confidence, reduce the risk of being suppressed by post-processing and be beneficial to subsequent text recognition.

To recognize the scene text, it is a strong requirement that the text can be tightly and robustly localized in advance. However, current datasets have very li{}ttle curve text, and it is defective to label such text with quadrangle let alone rectangle. For example, as showed in Figure~\ref{fig:background}, using curve bounding box has three remarkable advantages:
\begin{itemize}
  \item \textbf{Avoid needless overlap}. Because the text may appear in many ways, the traditional four points localization may not handle such elusive peculiarity well. As shown in Figure~\ref{fig:background} (a), quadrilateral bounding box can not avoid a mass of tanglesome overlap while curve bounding box can.
  \item \textbf{Less background noise}. As shown in Figure~\ref{fig:background} (b), if text appear in the curve form, quadrilateral bounding box suffers from background noise.
  \item \textbf{Avoid multiple text lines}. The recent popular recognition methods~\cite{shi2016robust,shi2016end,ablavatski2017enriched,lee2016recursive} all require single row text line in each bounding box. However, in some cases like Figure~\ref{fig:background} (c), quadrilateral bounding box can not avoid multiple text lines from disturbing with each others while curve bounding box can exquisitely solve this problem.
\end{itemize}

Actually, curve text are also very common in our real-world. For examples, text in most kinds of columnar objects (bottles, stone piles, etc.), spherical objects, plicated plane (clothes, streamer, etc.), coins, logos, signboard and so on. However, to our best knowledge, current methods can not directly detect the curve text. The linking methods~\cite{tian2016detecting,shi2017detecting,hu2017wordsup} can detect components of the text and then grouping them together to match the curve bounding box. But if there are many texts stack up together, in many cases like Figure~\ref{fig:background} (a), empirical connection rules are almost impossible to group the tiny components properly, and somehow these methods always bring more false positives than direct detection manners in practice.

\begin{figure*}[htb]
  \centering
  \centerline{\includegraphics[width=17cm, height = 10cm]{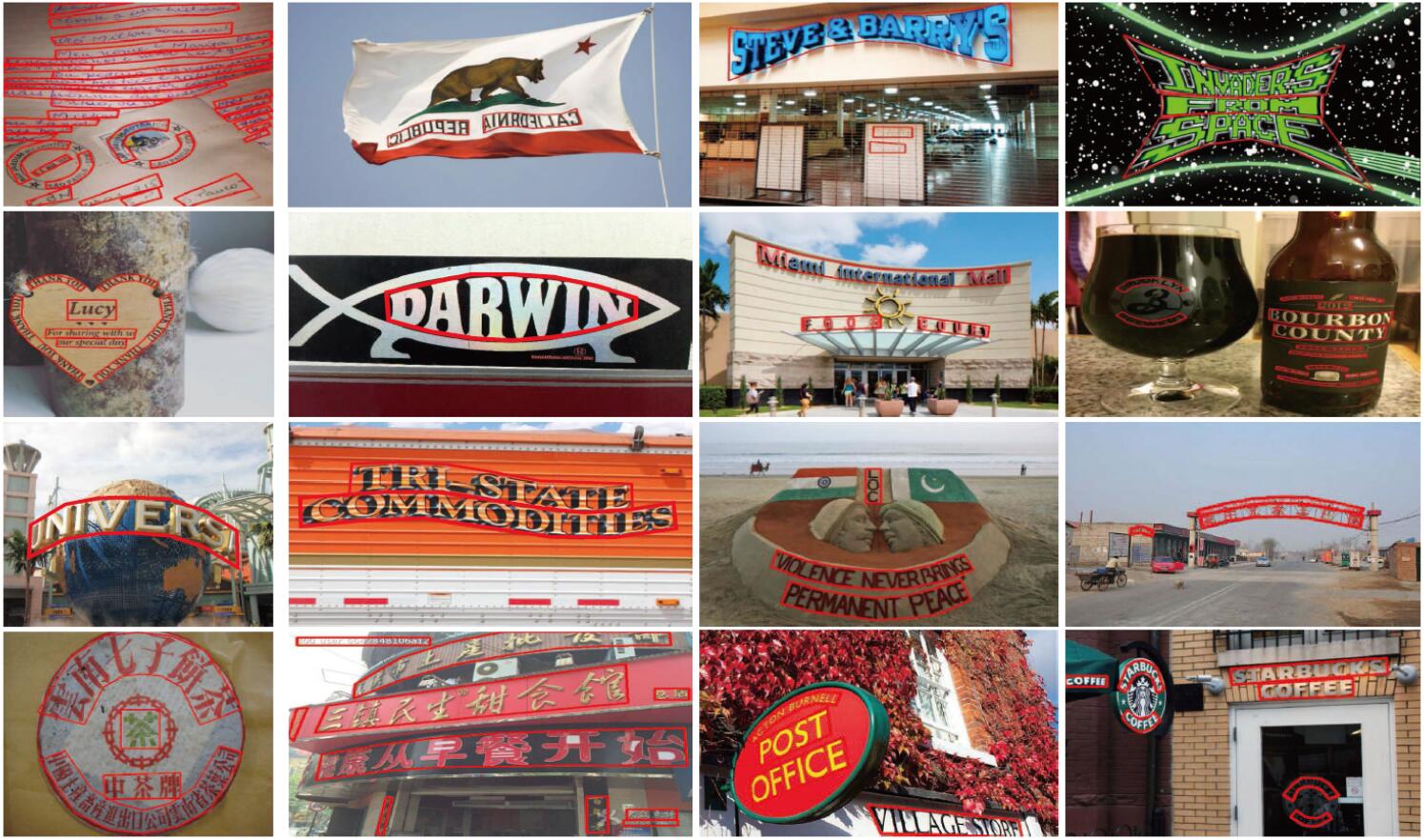}}
  \caption{Examples curve text annotations of CTW1500 dataset.}\label{fig:dataset}
\end{figure*}

Therefore, in this paper, we collect text from various natural scenes, websites or image library like google open-image \cite{openimages}, and constructing a new dataset named CTW1500. This dataset contains 1500 images with over 10k text annotation, and each image contains at least one curve text. For evaluation and comparison, we split 1000 images as training set and 500 for testing.
Based on our observation, for all kinds of the curve text regions, a 14 points polygon can be sufficient to localize them as shown in Figure~\ref{fig:background} and Figure~\ref{fig:dataset}. By using the referenced equant lines, it doesn't require much manpower to label as introduced in Sec. 3.

Based on the proposed dataset, we propose a simple but effective polygon based curve text detector (CTD), which may be a pioneering method that can directly detect the curve text. Unlike traditional detecting methods, the CTD separates the branches for width/height offsets predictions, which can be run below 4GB video memory with the speed of 13 FPS. In addition, the network architecture can be seamlessly integrated with a ingenious method we proposed, namely transverse and longitudinal offset connection (TLOC), which uses RNN to learn the inherent connection between locating points, making the detection more accurate and smooth. The CTD is also designed as an universal method, which can be trained with rectangular and quadrilateral bounding boxes without extra manual labels. Two simple but effective post-processing methods named non-polygon suppression (NPS) and polygonal non-maximum suppression (PNMS) are proposed to further intensify the generalization ability of CTD.

On the proposed dataset, the results demonstrate the CTD with a light reduced resnet-50 \cite{he2016deep} can effectively detect the curve text and outperform state-of-the-art methods with a large margin. By using TLOC, our method can remarkably improve the performance. Furthermore, we also evaluate our method on mere curve or non-curve test subset (by making the other text as not care regions), and the CTD+TLOC can still achieve the best results.

\section{Related Work}
In the past dozen years, scene text detection methods have achieved great improvement.
One of the main reasons of such unceasing progress is the evolution of the benchmark datasets - the data become harder, the amounts become larger and the labels become tighter. From 2003 on, rectangular labeled datasets such as ICDAR'03~\cite{lucas2003icdar}, ICDAR'11~\cite{shahab2011icdar}, ICDAR'13~\cite{Karatzas2013ICDAR} and COCO-Text~\cite{veit2016coco} have attracted a great number of research efforts. After 2010, the multi-oriented datasets with rotated rectangular labels (NEOCR \cite{nagy2011neocr}, OSTD \cite{yi2011text}, MSRA-TD500 \cite{Yao2012Detecting} and USTB-SV1K \cite{Yin2015Multi}) come out, which stimulate many influential multi-oriented detecting methods in the literatures. And in 2015, the first quadrilateral labeled dataset ICDAR 2015 ``Incidental Scene Text'' \cite{karatzas2015icdar} appears, which unprecedentedly attracted lots of attention according to its evaluating website \cite{karatzas2015icdar} and many recent progress.
 Since then, quite a few larger and more challenge quadrilateral labeled dataset in ICDAR 2017 competition like RCTW-17 \cite{Shi2017ICDAR2017} (dataset for Chinese and English text), DOST \cite{dost2017icdar} (scene texts observed by video in the real environment) and MLT \cite{mlt2017icdar} (dataset for multi-lingual text) seem to become the next mainstream datasets.

Interestingly, the development of detecting manners also show a similar evolution tendency with datasets.
Although the rectangular methods still receive interests, they seem becoming less mainstream. Since 2011, almost every year there are methods with rotated rectangular bounding box proposed. And in 2017, plentiful quadrilateral based detection methods~\cite{liu2017deep,shi2017detecting,zhou2017east,he2017deep,dai2017fused} emerged. Basically, the quadrilateral based detection methods can also achieve best performance in rotated datasets or horizontal datasets (by evaluating the circumscribed rectangle), and they all show a fact that they can beat horizontal manners in multi-oriented datasets (by using the circumscribed rectangle to train and test) especially in terms of recall rate. This is mainly because the stronger supervision for quadrilateral labeled methods can avoid much background noise, unreasonable suppression and information loss. The viewpoint that stronger supervision helps detection can also be found on Mask-RCNN~\cite{He2017Mask} which improves detecting results by jointly training with a branch of segmentation, and Ren et at. \cite{li2017towards} also shows training with recognition can be conducive to text detection.

However, current text detection methods, even quadrangle based methods all show disappointing performance in curve texts, which are commonly appeared in our real world as introduced in section 1. The linking method like~\cite{shi2017detecting} can not detect the strong bending text (the last image in second row in its Fig. 3) as well. One reason is that all current datasets contain very little curve text and many of the curve text are labeled with unsatisfactory rectangles. The other reason is that current four points based detection methods can only loosely detect the curve text, which may cause severely mutual interference like Figure~\ref{fig:background} (c). Therefore, in order to address the challenging problem of detecting curve text in the wild, we construct a new curve text based dataset named CTW1500, and then propose a novel method that can directly detect the curve text effectively.

\section{CTW1500 Dataset and Annotation}

{\bf{Data description}}. The CTW1500 dataset contains 1500 images, with 10,751 bounding boxes (3,530 are curve bounding boxes) and at least one curve text per image. The images are manually harvested from internet, image library like google Open-Image \cite{openimages} and our own data collected by phone cameras, which also contain lots of horizontal and multi-oriented text. The distribution of the images is various, containing indoor, outdoor, born digital, blurred, perspective distortion texts and so on. In addition, our dataset is multi-lingual with mainly Chinese and English text.
\begin{figure}[htb]
  \centering
  \centerline{\includegraphics[width=8.5cm, height = 4.5cm]{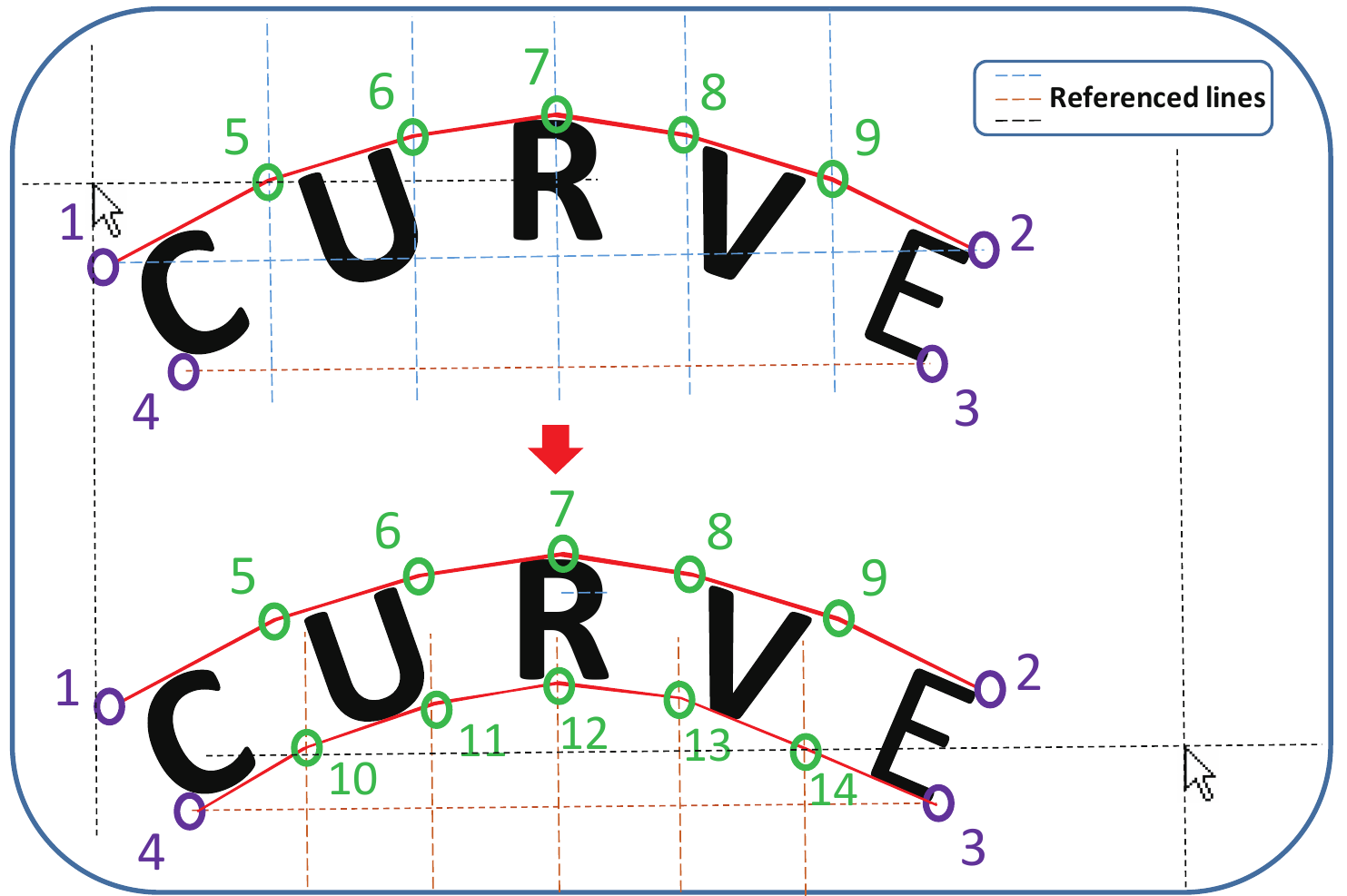}}
  \caption{Illustration of labeling curve text.}\label{fig:label}
\end{figure}

{\bf{Annotation}.} The text is manually labeled by ourself with our labeling tool. For labeling text with horizontal or quadrilateral shape, simply two or four clicks are required. To surround the curve text, we create ten equidistant referenced lines to help label the extra 10 points (we practically find extra 10 points are sufficient to label all kinds of curve text as shown in Figure \ref{fig:dataset}). The reason we use the equidistant lines is to ease the labeling effort, and reduce subjective interference.
To evaluate the localization performance, we simply follow the PASCAL VOC protocol \cite{everingham2010pascal}, which uses 0.5 IoU threshold to decide true or false positive. The only difference is we calculate the exact intersection-over-union (IoU) between the polygons instead of axis-aligned rectangles.

The labeling procedure is shown in Figure \ref{fig:label}. Firstly, we click the four vertexes marked as 1, 2, 3, 4, and the referenced dashed line (blue) will be automatically created. Moving one of the mouse's referenced line (horizontal and vertical black dashed lines) to the appropriate position (intersection of two referenced lines) and then click to determine the next point, and so on for the remanent points. We roughly calculate the labeling time of three shapes of text in Table \ref{tab:label}, which shows labeling one curve text consumes approximately triple time than labeling with quadrangle. The CTW1500 dataset can be download at {\color{blue} https://github.com/Yuliang-Liu/Curve-Text-Detector}.

\begin{table}[!t]
\caption{Time cost of labeling different shapes of text.}
\label{tab:label}
\centering
\small
\begin{tabular}{|c|c|c|c|}
  \hline
  Bounding Box & Horizontal  & Quadrilateral & Curve \\
  \hline
  \hline
   Labeling Time (s) & 2.5 & 4 & 13 \\
   \hline
\end{tabular}
\end{table}

\section{Methodology}
This section presents details of our curve text detector (CTD). We will first illustrate the architecture of the CTD and how we make use of the polygonal labels. After that, we will describe how a recurrent neural network (RNN) component is seamlessly connected to the CTD and subsequently introducing the universality of this method. Finally, we will present our two simple but effective post-processing methods that can further improve the performance. 

\subsection{Network Architecture}
\begin{figure*}[htb]
  \centering
  \centerline{\includegraphics[width=17cm, height = 9cm]{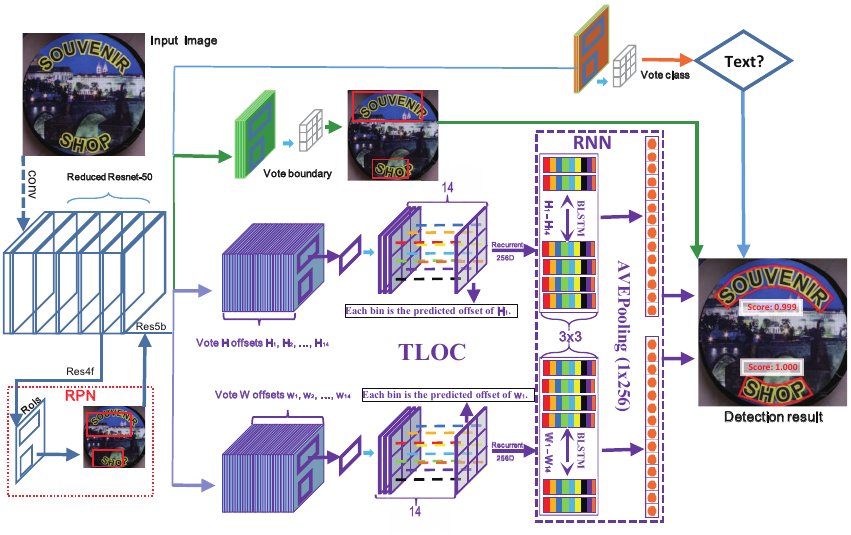}} 
  \caption{Overall structure of our Curve Text Detector (CTD).  }\label{fig:net}
\end{figure*}

The overall architecture of our CTD is shown in Figure ~\ref{fig:net}, which can be divided into three parts: backbone, RPN and regression module. Backbone usually adopts the popular models pre-trained by ImageNet \cite{deng2009imagenet} and then uses the corresponding model to finetune, such as VGG-16 \cite{simonyan2014very}, ResNet \cite{he2016deep} and so on. Region proposal network (RPN) and regression module are respectively connected to the backbone; while the former generates proposals for roughly recalling text and the latter finely adjusts the proposals to make it tighter.

In this paper, we use a reduced ResNet-50 (simply remove the last residual block) as our backbone, which requires less memory and can be faster. In the RPN stage, we use the default rectangular anchors to roughly recall the text but we set a very loose RPN-NMS threshold to avoid premature suppression.
To detect the curve text with polygon, the CTD only need to modify the regression module by adding the curve locating points, which is inspired by DMPNet \cite{liu2017deep} and East \cite{zhou2017east} that adopting quadrilateral regressing branch separated with circumscribed rectangle regression. The rectangular branch can be easily learned by the network and let it converses fast, which can also roughly detect the text region in advanced and alleviate the following regression. Contrastively, the quadrilateral branch offers stronger supervision to guide the network being more accurate.

Similar to \cite{ren2015faster,liu2017deep}, we also regress the relative positions for each points. Unlike \cite{liu2017deep}, we use the minimum x and minimum y of the circumscribed rectangle as the datum point. Therefore, the relative length $w_i$ and $h_i$ ($i$ $\in$ $1,2,...,14$) of every point is greater than zero, which is somehow easier to train in practice. In addition, we separately predict the offsets $w$ and $h$, which can not only reduce the parameters but more reasonable for sequential learning as introduced in the following subsection.
The total number of the regressing items is 32; while 28 are the offset of the 14 points, and 4 are the x,y minimum and maximum of the circumscribed rectangle. The parameterizations of the 14 offsets ($d_{w_{i}}$ and $d_{h_{i}}$) are listed below:
\begin{equation}\label{eq:regress}
  \left\{
          \begin{array}{ll}
            d_{w_{i}} = \frac{p^{*}_{w_{i}}-p_{w_{i}}}{w_{chr}}, \\
            d_{h_{i}} = \frac{p^{*}_{h_{i}}-p_{h_{i}}}{h_{chr}},
          \end{array}
        \right. (i\in (1,2,...,14))
\end{equation}
Where, $p^{*}$ and $p$ are ground truth and predicted offsets respectively. Besides, $w_{chr}$ and $h_{chr}$ are the width and height of the circumscribed rectangle. For boundary regression, we follow the same as Faster R-CNN \cite{ren2015faster}. It is worth noticing that 28 values are enough to determine the position of 14 points, but in relative regressing mode, 32 values can be easier to retrieve the rest of 14 points and offer stronger supervision.

\subsection{Recurrent Transverse and Longitudinal Offset Connection (TLOC)}
Using recurrent connection in text detection task was proved robust and effective by CTPN \cite{tian2016detecting}, which learns the latent sequence of tiny proposals and produces better results. However, CTPN is a linking based method, which requires empirical connection. Also, its connectionist proposal requires fixed images size for ensuring the fixed number of input time sequences of RNN.
Unlike CTPN, our method can directly localize the curve region without exterior connection, and the number of the time sequences of RNN is not constrained by input image size. This is because the RNN is connected to the output of the position-sensitive RoI Pooling (PSROIPooling) \cite{dai2016r}, and the number of the output targets are fixed (14 width offsets and 14 height offsets).
Basically, PSROIPooling is used to predict and vote the class probabilities and localization offsets, which evenly partitions each RoI into $p\times p$ bins to estimate position information. The dimension of the input convolutional layer should be $(class+1)p^2$, and thus the PSROIPooling can produce a $p^2$ score map for each category. For transverse and longitudinal offset prediction, we remove the background class localization score map and use a $7\times 7$ bin, so the input convolutional dimension is $14\times 7\times 7$ for height and width separately. Each value of $(i; j)$-th bin $(0 < i; j < p-1)$ is computed from the corresponding position in the (i; j)-th score map by using a average pooling:
\begin{equation}\label{eq:psroi}
  r_{c}(i, j|\Theta)|= \sum\limits_{(x,y)\in bin(i,j)}\frac{s_{i,j,c}(x+x_{min}, y+y_{min}|\Theta)}{n},
\end{equation}
where, $r_c(i,j|\Theta)$ is the pooled value in the $(i,j)$-th bin for category $c$, $s_{i,j,c}$ represents a score map from the corresponding dimension. $(x_{min},y_{min})$ denotes the left-top coordinate of a RoI, $n$ denotes the amount of pixels in the bin, and $\Theta$ stands for all network parameters. After the PSROIPooling procedure, CTD will get the scores or the estimated offsets of each RoI via globally pooling on the $p^2$ position-sensitive score maps: $r_c(\Theta) = \sum\frac{r_c(i,j|\Theta)}{p^2}$, which produces a $(C + 1)$-dimensional vector.
The voting class score is then computed by the softmax operation across all categories and output the final confidence:
\begin{equation}\label{eq:softmax}
  s_{c}(\Theta) = e^{r_{c}(\Theta)}\diagup \sum\limits_{c^{'}=0}^{C}e^{r_{c^{'}}(\Theta)}.
\end{equation}
The localization offsets would be fed into the localization loss function. During training phase, we choose the similar multi-task loss functions for score and offsets prediction as follow:
\begin{equation}\label{eq:optimazation}
\begin{aligned}
  L(c,c^{*},b,b^{*},w,w^{*},h,h^{*}) = \frac{1}{N}(\lambda\times L_{cls}(c, c^{*}) + \\
  L_{loc}(b,b^{*})) + \frac{\mu}{N_{p}}(L_{loc}(h,h^{*}) + L_{loc}(w,w^{*}))
\end{aligned}
\end{equation}
where $N$ is the amount of both positive and negative proposals that match specific overlapping range and $N_{p}$ is the number of positive proposals because it is not necessary to refine the negative proposals. Besides, $\lambda$ and $\mu$ are balance factors which weighs the importance among the classification and detection losses ($L_{cls}$ represents classification loss function; $L_{loc}$ is localization loss function which can be smooth-$L1$ loss or smooth-$Ln$\cite{liu2017deep} loss). Practically, we set $\lambda$ to 3 or even more to balance localization loss which has much more targets. Moreover, $(c,b,w,h)$ represent the predicted class, estimated bounding-box, width and height offset respectively, and $(c^{*},b^{*},w^{*},h^{*})$ denote the corresponding ground-truth.

\begin{figure}[htb]
  \centering
  \centerline{\includegraphics[width=8cm, height = 3cm]{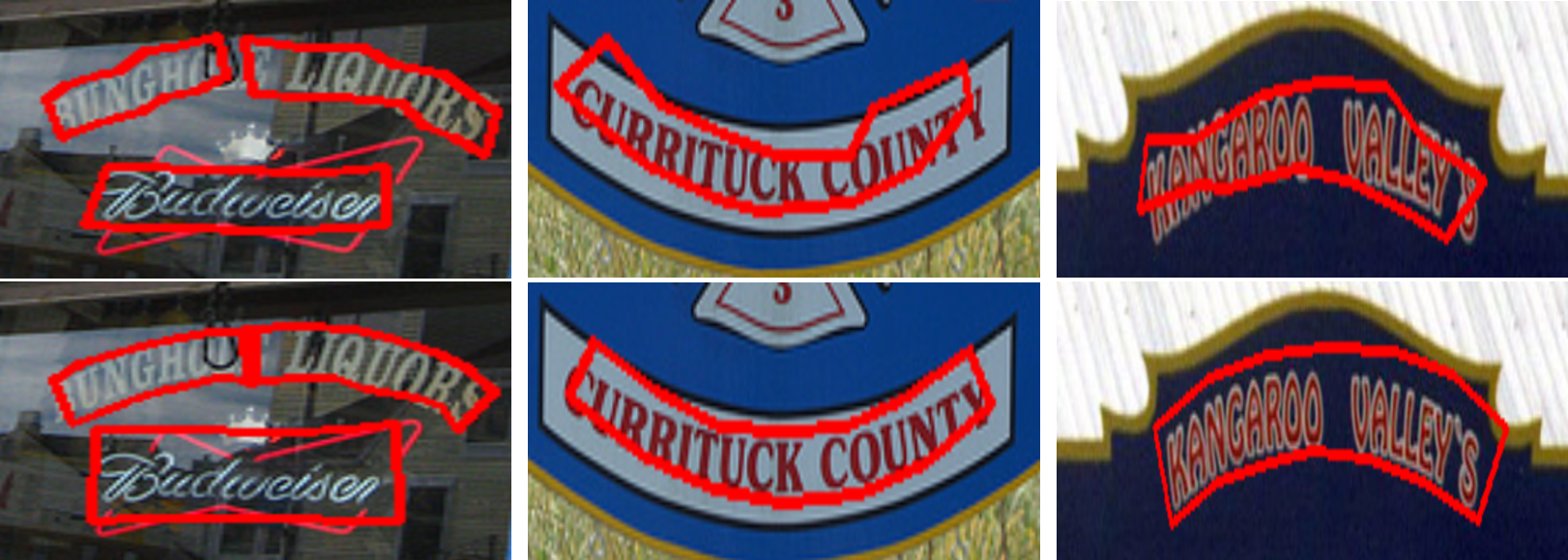}} 
  \caption{Top: Detection results without TLOC. Bottom: Detection results with TLOC.}\label{fig:smooth}
\end{figure}

To improve detection performance, we separate transverse and longitudinal branches to predict the offsets for localizing the text region. Intuitively, each point is restricted by the last and next points and the textual region. For example, in the case of Figure \ref{fig:label}, the offset width of the sixth labeling point should larger than the fifth point and less than the seventh point. Independently predicting each offset may lead to unsmooth text region, and somehow it may bring more false detection. Therefore, we assume the width/height of each point has associated context information, and using RNN to learn their latent characteristics. We name this method as recurrent transverse and longitudinal offset connection (TLOC). The TLOC structure is shown with purple patterns in Figure \ref{fig:net} and we also list some examples in Figure \ref{fig:smooth} to show the difference of whether adding TLOC or not.
To adopt TLOC, we find the output of PSROIPooling is suitable to encode the offsets context information. Take the width offset branch as an example; firstly, PSROIPooling outputs 14 $p^2$ score map for voting $w_1, ..., w_{14}$ of each proposal, and $p^2$ bins of the $i$-th score map have $p^2$ voting values from respective position, which can be encoded as the feature of $w_i$; The RNN than takes width offsets feature of each point as sequential inputs, and recurrently updates the inherent state inside the hidden layers, $L_t$, i.e.
\begin{equation}\label{eq:recur}
  L_{t} = \varphi(L_{t-1}, O_t), t = 1,2,...,14
\end{equation}
where $O_t \in \Re^{P^{2}}$ is the $t$-th predicting offset from the corresponding psroi-pooling output channel. $L_t$ is a recurrent internal state computed from both current input ($O_t$) and the last state encoded in $L_{t-1}$. The recurrence is computed by using a non-linear function $\varphi$, where we adopt a bi-directional long short-term memory (BLSTM) architecture \cite{hochreiter1997long} as our RNN. The internal state inside the RNN hidden layer associates the sequential context information by all previous estimated offsets through the recurrent connection, and we empirically use a 256D BLSTM hidden layer, thus $L_{t} \in \Re^{256}$. Finally, the output of the BLSTM is a $14$ dimensional $1\times 256$ vector, which is globally pooled by a ($1\times 256$) kernel to output the final prediction.

\subsection{Long Side Interpolation}
\begin{figure}[htb]
  \centering
  \centerline{\includegraphics[width=8cm, height = 3.2cm]{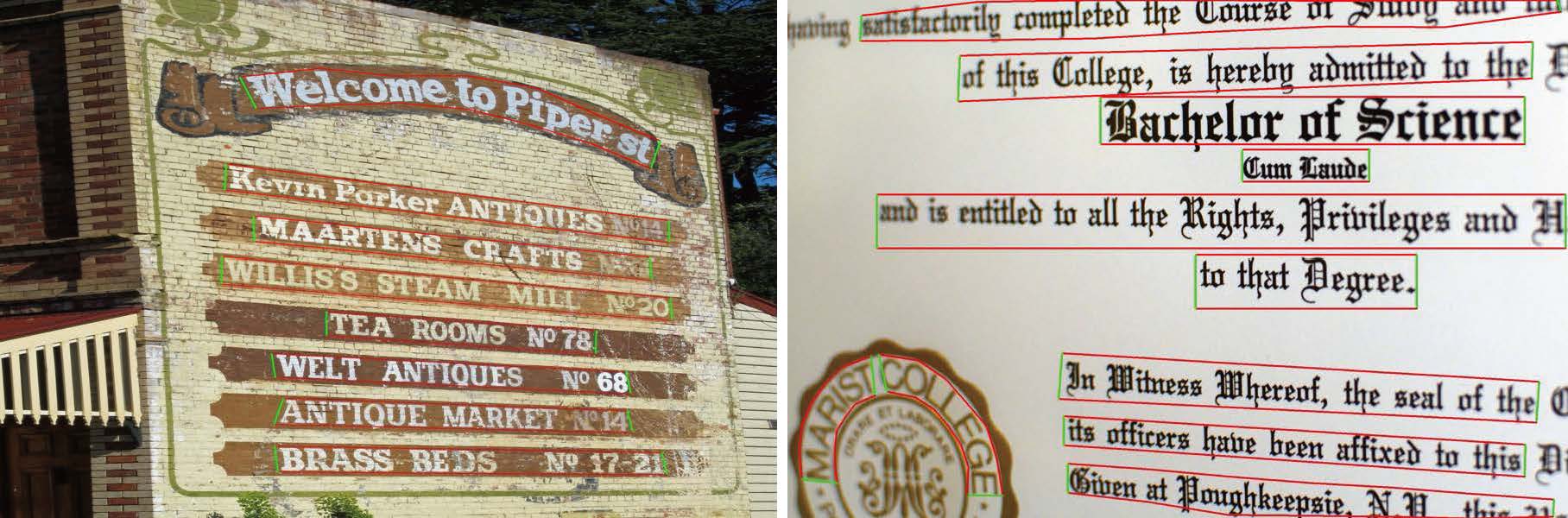}}
  \caption{Visualization of the interpolation for 4 points bounding boxes. The 10 equal division points will be respectively interpolated in two {\bf{Red}} sides of each bounding box. {\bf{Green}} means straight line without interpolation.}\label{fig:interpo}
\end{figure}

As introduced in Sec. 4.1, for each bounding box, the CTD outputs offsets for 14 points. However, almost all current benchmark datasets only have labels of two or four vertexes information. To train in these datasets, we can easily interpolate the equal division points in the largest side and its opposite side as shown in \ref{fig:interpo}.

\begin{figure*}[htb]
  \centering
  \centerline{\includegraphics[width=17cm, height = 8.4cm]{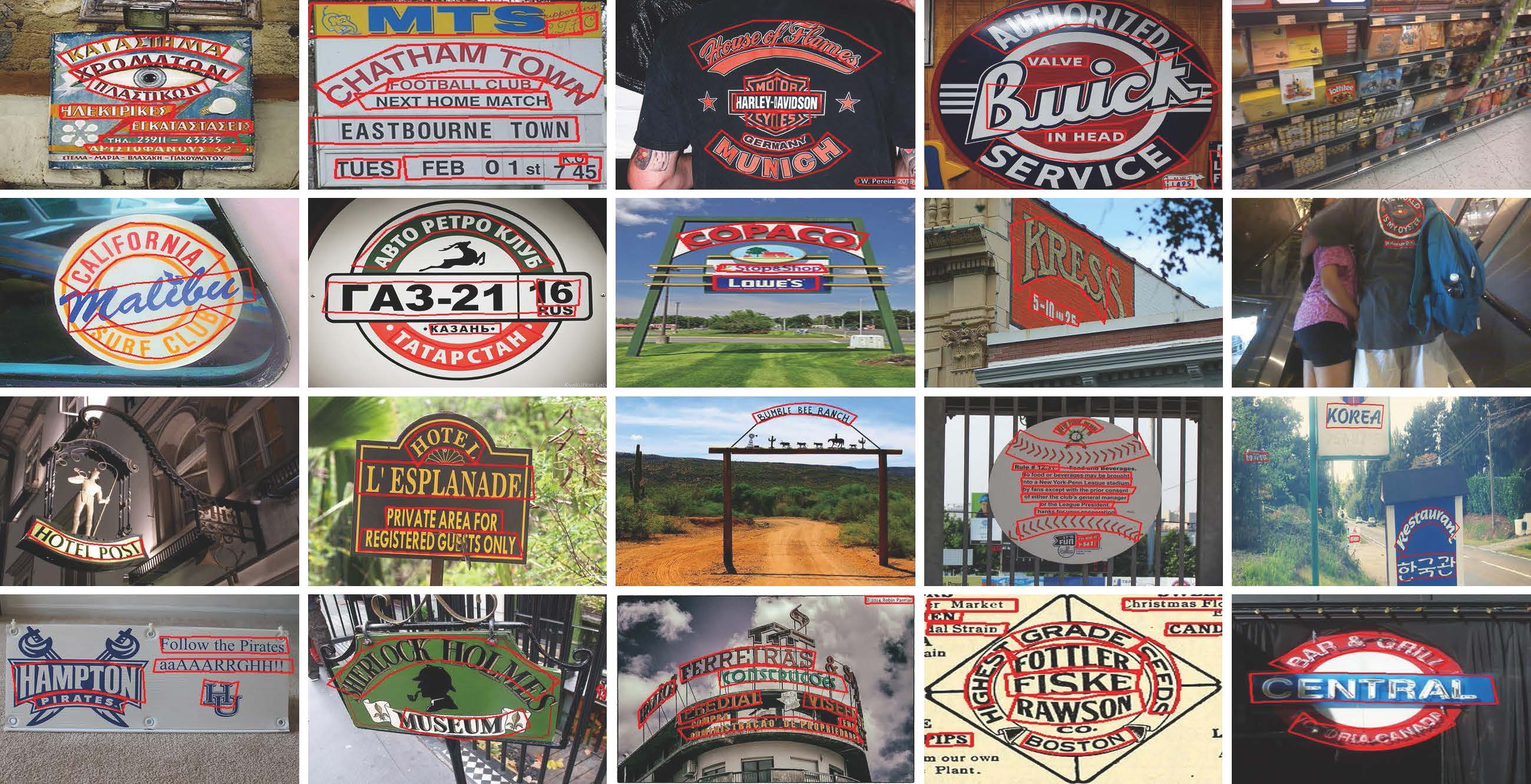}}
  \caption{Detection results visualization. The fourth column lists some inferior results and the images from last column are from other datasets for further testing the generalization ability.}\label{fig:det:oth:cur}
\end{figure*}

Promisingly, by simply interpolating the points in the bounding box, our CTD can be effectively trained with all text region and it can achieve better results as demonstrated in the experiment section.

\subsection{Polygonal Post Processing}
{\bf{Non-polygon suppression (NPS)}}.
False positive detecting results are one of the important reasons that restricting the performance of text detection. However, in CTD, some diffident false positives will appear with invalid shape (for a valid polygon, there is not any intersecting side). In addition, there is hardly any scene text come out with intersecting side, and these invalid polygons are nearly impossible to recognize. Therefore, we simply suppress all these invalid polygons and we named it a non-polygon suppression (NPS), which can slightly improve the accuracy without influencing the recall rate.

{\bf{Polygonal non-maximum suppression (PNMS)}}.
Non-maximum suppression \cite{Neubeck2006Efficient} is proved very effective for object detection task. Because of the particularity of the curve scene text, rectangular NMS is limited to handle dense multi-oriented text as shown in Figure \ref{fig:background} (a) and (c). To solve this problem, \cite{zhou2017east} propose a locality-aware NMS and \cite{dai2017fused} devise a Mask-NMS to suppress the final output results. In this paper, we also improve the NMS by computing the overlapping area between polygons, named polygon non-maximum suppression (PNMS), which is proved effective in the following experiments.

\begin{table*}[!t]
\caption{Experiments on the proposed CTW1500 test set, curve subset and non-curve subset. R: Recall. P: Precision. H: Hmean. S: Speed. Speed column evaluates the time of the forward procedure without post-processing. }
\label{tab:cur}
\centering
\begin{tabular}{ccccccccccccc}
  \toprule
  \multirow{2}*{{\bf{Algorithm}}} &  \multicolumn{3}{c}{{\bf{Whole set}}} & & \multicolumn{3}{c}{{\bf{Non-curve subset}}} &   &  \multicolumn{3}{c}{{\bf{Curve subset}}}& \multirow{2}*{{\bf{S (FPS)}}} \\
  \cline{2-4}
  \cline{6-8}
  \cline{10-12}
                &   R (\%)  & P (\%) & H (\%) &  &   R (\%)  & P (\%) & H (\%) &   & R (\%)  & P (\%) & H (\%) & \\
  \bottomrule
  Seglink \cite{shi2017detecting} & 40.0 & 42.3 & 40.8 & & 48.4 & 38.3 & 42.8 & & 19.4 & 9.9 & 13.2 & 10.7\\
  \hline
  SWT \cite{epshtein2010detecting} & 9.0 & 20.7 & 12.5 & & 5.8 & 13.4 & 8.1 & & 6.4 & 7.0 & 6.7 & - \\
  \hline
  CTPN \cite{tian2016detecting} & 53.8 & 60.4 & 56.9 &  & 59.4 & 54.3 & 56.7 & & 37.7 & 34.1 & 35.8 & 7.14\\
  \hline
  EAST \cite{zhou2017east} & 49.1 & 78.7 & 60.4 & & 57.5 & 71.0 & 63.6 & & 29.9 & 40.9 & 34.6 & {\bf{21.2}}\\
  \hline
  DMPNet \cite{liu2017deep} & 56.0 & 69.9 & 62.2 &  & 61.7 & 63.9 & 62.7 & & 39.3 & 35.5 & 37.3 & 12.3 \\
  \hline
  AdaBoost \cite{chen2004detecting} & 4.4 & 6.7 & 5.3 &   & - & - & - &  & - & - & - & - \\
  \midrule
  \midrule
  CTD (ours) & 65.2 & 74.3 & 69.5 & & 60.3 & 67.3 & 63.5 & & 73.9 & 52.9 & 61.6 & 15.2\\
  \hline
  CTD + TLOC (ours)  & 69.8 & 77.4 & {\bf{73.4}} & & 62.3 & 70.8 & {\bf{66.3}} & & 77.1 & 57.1 & {\bf{65.6}} & 13.3\\
  \bottomrule
  \end{tabular}
\end{table*}

\begin{table}[!t]
\caption{Experiments to evaluate TLOC and PNMS. Here we can use NMS is because CTD remains the circumscribed rectangle branch, and the float number is the threshold. R: Recall rate. P: Precision. H: Hmean. }
\label{tab:tloc}
\centering
\normalsize
\begin{tabular}{cccc}
  \toprule
  Algorithm & R (\%)  & P (\%) & H (\%) \\
  \bottomrule
  CTD + NMS0.3 & 64.4 & 74.9 & 69.3 \\
  \hline
  CTD + PNMS0.1 & 65.2 & 74.3 & 69.5 \\
  \hline
  CTD + TLOC + PNMS0.1 & 69.8 & 77.4 & {\bf{73.4}} \\
  \hline
  CTD + TLOC + PNMS0.2 & 70.1 & 75.0 & 72.4 \\
  \hline
  CTD + TLOC + PNMS0.3 & 70.8 & 71.6 & 71.2 \\
  \hline
  CTD + TLOC + PNMS0.4 & 71.7 & 65.3 & 68.3 \\
  \hline
  CTD + TLOC + NMS0.1 & 63.7 & 78.4 & 70.3 \\
  \hline
  CTD + TLOC + NMS0.2 & 68.6 & 78.1 & 73.1 \\
  \hline
  CTD + TLOC + NMS0.3 & 69.7 & 77.1 & 73.2 \\
  \hline
  CTD + TLOC + NMS0.4 & 70.8 & 74.7 & 72.7 \\
  \bottomrule
\end{tabular}
\end{table}

\section{Experiments}
In this section, we carry experiments on the proposed CTW1500 dataset to test our method.
The testing environment is Ubuntu 16.04 64bit with single Nvidia 1080 GPU. All the text detection results are evaluated with the protocol introduced in section 3, which exactly calculates the IoU between the polygons instead of axis-align rectangles. For fair comparison, all the experiments only use the provided training data without any data augmentation. 

{\bf{Effectiveness of TLOC and PNMS}}. We first evaluate the availability of proposed TLOC and PNMS, and the results are given in Table ~\ref{tab:tloc}. In this table, we can find whatever using CTD or CTD + TLOC, the PNMS can always slightly outperform the classic NMS. On the other hand, the results also demonstrate that by simply adding TLOC, the proposed CTD can be improved about 4 percents in terms of Hmean. Note that we don't compare NPS here because it is a prerequisite post-processing method for evaluating polygonal overlapping area, which can slightly improve the accuracy.

{\bf{Comparison with state-of-the-arts methods}}. For comprehensively evaluating our method, we compare the proposed CTD and CTD + TLOC with several state-of-the-art and common text detection methods. Note that for East \cite{zhou2017east} and Seglink \cite{shi2017detecting}, we re-implement these methods based on the unofficial source codes from github, and for fair comparison, we do not use huge synthetic data to pre-train seglink which may reduce its performance. For DMPNet \cite{liu2017deep} and CTPN \cite{tian2016detecting}, we both use Caffe \cite{jia2014caffe} framework to re-implement these methods. Besides, because none of these methods can be trained with curve text region, we follow the traditional circumscribed rectangular bounding box to make the labels trainable for them. Table \ref{tab:cur} lists the experimental results. The results of the whole CTW-1500 test set show that the proposed CTD + TLOC can outperform state-of-the-art methods with more than 10 percents in terms of Hmean. To further evaluating the effectiveness, we also split the curve and non-curve text from the whole test set by simply regarding the other kind of text as difficult (not care) and making comparison. Promisingly, in the curve subset, proposed CTD + TLOC can outperform state-of-the-art methods with at least 28 percents of Hmean, and meanwhile, this method can also achieve the best results in detecting only non-curve texts, demonstrating its robustness and universality. Note that for our method, the results of the subset are both less than the whole set are because the way treating the other text as difficult will reduce the precision.

Besides, we compare the detection speed in the last column of the Table \ref{tab:cur} and the results (13.3 or 15.2 FPS) show that our method is the second fastest, which also proves the effectiveness of our method.  

Examples of the detection results are visualized in Figure \ref{fig:det:oth:cur}. In the last column of this figure, we also use our CTD to detect the curve texts from other dataset, which qualitatively demonstrates its powerful ability for detecting curve texts, as well as its generalization ability.

\section{Conclusions and Future Work}
Curve text is common in our real world but currently few datasets or methods are aiming at curve text detection. In order to facilitate this new challenging research of reading curve text in the wild, in this paper, we propose a new dataset named CTW1500, which is a novel dataset mainly constructed by curve text. The curve text in this dataset are approximated labeled by polygon which do not require too much manpower. In addition, we propose a new CTD approach which may be the first attempt to directly detect the curve texts. By devising a transverse and longitudinal offset connection (TLOC) method, the CTD can be seamlessly connected with RNN, which significantly improves the detecting performance.
We also propose a simple but effective long side interpolation technique, which let CTD become an universal method that can also be trained with rectangular or quadrilateral bounding boxes without additional manual efforts. Lastly, we design two post-processing methods that are also demonstrated effective.

In future, the propose dataset could be enlarged as a curve text based recognition dataset, because the labeling manner seems good for recognition. Besides, although the flexible detecting manner like our CTD  may be slightly slower than an rigid rectangle based detector, the former can solve more complicated problems like detecting curve text and achieving better results, which should worth further exploration.

{\small
\bibliographystyle{ieee}
\bibliography{refs}

\begin{thebibliography}{10}\itemsep=-1pt

\bibitem{ablavatski2017enriched}
A.~Ablavatski, S.~Lu, and J.~Cai.
\newblock Enriched deep recurrent visual attention model for multiple object
  recognition.
\newblock In {\em Applications of Computer Vision (WACV), 2017 IEEE Winter
  Conference on}, pages 971--978. IEEE, 2017.

\bibitem{chen2004detecting}
X.~Chen and A.~L. Yuille.
\newblock Detecting and reading text in natural scenes.
\newblock In {\em Computer Vision and Pattern Recognition, 2004. CVPR 2004.
  Proceedings of the 2004 IEEE Computer Society Conference on}, volume~2, pages
  II--II. IEEE, 2004.

\bibitem{dai2016r}
J.~Dai, Y.~Li, K.~He, and J.~Sun.
\newblock R-fcn: Object detection via region-based fully convolutional
  networks.
\newblock In {\em Advances in neural information processing systems}, pages
  379--387, 2016.

\bibitem{dai2017fused}
Y.~Dai, Z.~Huang, Y.~Gao, and K.~Chen.
\newblock Fused text segmentation networks for multi-oriented scene text
  detection.
\newblock {\em arXiv preprint arXiv:1709.03272}, 2017.

\bibitem{deng2009imagenet}
J.~Deng, W.~Dong, R.~Socher, L.-J. Li, K.~Li, and L.~Fei-Fei.
\newblock Imagenet: A large-scale hierarchical image database.
\newblock In {\em Computer Vision and Pattern Recognition, 2009. CVPR 2009.
  IEEE Conference on}, pages 248--255. IEEE, 2009.

\bibitem{epshtein2010detecting}
B.~Epshtein, E.~Ofek, and Y.~Wexler.
\newblock Detecting text in natural scenes with stroke width transform.
\newblock In {\em Computer Vision and Pattern Recognition (CVPR), 2010 IEEE
  Conference on}, pages 2963--2970. IEEE, 2010.

\bibitem{everingham2010pascal}
M.~Everingham, L.~Van~Gool, C.~K. Williams, J.~Winn, and A.~Zisserman.
\newblock The pascal visual object classes (voc) challenge.
\newblock {\em International journal of computer vision}, 88(2):303--338, 2010.

\bibitem{He2017Mask}
K.~He, G.~Gkioxari, P.~Dollár, and R.~Girshick.
\newblock Mask r-cnn.
\newblock {\em Proceedings of the IEEE International Conference on Computer
  Vision}, 2017.

\bibitem{he2016deep}
K.~He, X.~Zhang, S.~Ren, and J.~Sun.
\newblock Deep residual learning for image recognition.
\newblock In {\em Proceedings of the IEEE conference on computer vision and
  pattern recognition}, pages 770--778, 2016.

\bibitem{he2017deep}
W.~He, X.-Y. Zhang, F.~Yin, and C.-L. Liu.
\newblock Deep direct regression for multi-oriented scene text detection.
\newblock {\em Proceedings of the IEEE International Conference on Computer
  Vision}, 2017.

\bibitem{hochreiter1997long}
S.~Hochreiter and J.~Schmidhuber.
\newblock Long short-term memory.
\newblock {\em Neural Computation}, 9(8):1735--1780, 1997.

\bibitem{hu2017wordsup}
H.~Hu, C.~Zhang, Y.~Luo, Y.~Wang, J.~Han, and E.~Ding.
\newblock Wordsup: Exploiting word annotations for character based text
  detection.
\newblock {\em Proceedings of the IEEE International Conference on Computer
  Vision}, 2017.

\bibitem{mlt2017icdar}
e.~a. I, Bizid.
\newblock ``icdar2017 robust reading competition – challenge on multi-lingual
  scene text detection and script identification'' [online].
\newblock {\em http://rrc.cvc.uab.es/?ch=8}, 2017.

\bibitem{dost2017icdar}
M.~T. M. N. S. H. I. Y. K. K. e.~a. Iwamura, M.
\newblock ``icdar 2017 robust reading challenge on omnidirectional video''
  [online].
\newblock {\em http://rrc.cvc.uab.es/?ch=7}, 2017.

\bibitem{jia2014caffe}
Y.~Jia and e.~a. Shelhamer, Evan.
\newblock Caffe: Convolutional architecture for fast feature embedding.
\newblock In {\em Proceedings of the 22nd ACM international conference on
  Multimedia}, pages 675--678. ACM, 2014.

\bibitem{karatzas2015icdar}
D.~Karatzas and e.~a. Gomez-Bigorda, Lluis.
\newblock Icdar 2015 competition on robust reading.
\newblock In {\em Document Analysis and Recognition (ICDAR), 2015 13th
  International Conference on}, pages 1156--1160. IEEE, 2015.

\bibitem{Karatzas2013ICDAR}
D.~Karatzas, F.~Shafait, S.~Uchida, M.~Iwamura, L.~G.~I. Bigorda, S.~R. Mestre,
  J.~Mas, D.~F. Mota, J.~A. Almazàn, and L.~P. D.~L. Heras.
\newblock Icdar 2013 robust reading competition.
\newblock In {\em International Conference on Document Analysis and
  Recognition}, pages 1484--1493, 2013.

\bibitem{openimages}
I.~Krasin and e.~a. Duerig, Tom.
\newblock Openimages: A public dataset for large-scale multi-label and
  multi-class image classification.
\newblock {\em Dataset available from https://github.com/openimages}, 2017.

\bibitem{lee2016recursive}
C.-Y. Lee and S.~Osindero.
\newblock Recursive recurrent nets with attention modeling for ocr in the wild.
\newblock In {\em Proceedings of the IEEE Conference on Computer Vision and
  Pattern Recognition}, pages 2231--2239, 2016.

\bibitem{li2017towards}
H.~Li, P.~Wang, and C.~Shen.
\newblock Towards end-to-end text spotting with convolutional recurrent neural
  networks.
\newblock {\em arXiv preprint arXiv:1707.03985}, 2017.

\bibitem{liu2017deep}
Y.~Liu and L.~Jin.
\newblock Deep matching prior network: Toward tighter multi-oriented text
  detection.
\newblock {\em Proceedings of the IEEE Conference on Computer Vision and
  Pattern Recognition}, 2017.

\bibitem{lucas2003icdar}
S.~M. Lucas, A.~Panaretos, L.~Sosa, A.~Tang, S.~Wong, and R.~Young.
\newblock Icdar 2003 robust reading competitions.
\newblock In {\em Document Analysis and Recognition, 2003. Proceedings. Seventh
  International Conference on}, pages 682--687. IEEE, 2003.

\bibitem{nagy2011neocr}
R.~Nagy, A.~Dicker, and K.~Meyer-Wegener.
\newblock Neocr: A configurable dataset for natural image text recognition.
\newblock In {\em International Workshop on Camera-Based Document Analysis and
  Recognition}, pages 150--163. Springer, 2011.

\bibitem{Neubeck2006Efficient}
A.~Neubeck and L.~V. Gool.
\newblock Efficient non-maximum suppression.
\newblock In {\em International Conference on Pattern Recognition}, pages
  850--855, 2006.

\bibitem{ren2015faster}
S.~Ren, K.~He, R.~Girshick, and J.~Sun.
\newblock Faster r-cnn: Towards real-time object detection with region proposal
  networks.
\newblock In {\em Advances in neural information processing systems}, pages
  91--99, 2015.

\bibitem{shahab2011icdar}
A.~Shahab, F.~Shafait, and A.~Dengel.
\newblock Icdar 2011 robust reading competition challenge 2: Reading text in
  scene images.
\newblock In {\em Document Analysis and Recognition (ICDAR), 2011 International
  Conference on}, pages 1491--1496. IEEE, 2011.

\bibitem{shi2017detecting}
B.~Shi, X.~Bai, and S.~Belongie.
\newblock Detecting oriented text in natural images by linking segments.
\newblock {\em Proceedings of the IEEE Conference on Computer Vision and
  Pattern Recognition}, 2017.

\bibitem{shi2016end}
B.~Shi, X.~Bai, and C.~Yao.
\newblock An end-to-end trainable neural network for image-based sequence
  recognition and its application to scene text recognition.
\newblock {\em IEEE transactions on pattern analysis and machine intelligence},
  2016.

\bibitem{shi2016robust}
B.~Shi, X.~Wang, P.~Lyu, C.~Yao, and X.~Bai.
\newblock Robust scene text recognition with automatic rectification.
\newblock In {\em Proceedings of the IEEE Conference on Computer Vision and
  Pattern Recognition}, pages 4168--4176, 2016.

\bibitem{Shi2017ICDAR2017}
B.~Shi, Yao, M.~Liao, Y.~M., X.~P., L.~Cui, L.~S. Serge~Belongie, and B.~X.
\newblock Icdar2017 competition on reading chinese text in the wild (rctw-17).
\newblock {\em arXiv preprint arXiv:1708.09585}, 2017.

\bibitem{simonyan2014very}
K.~Simonyan and A.~Zisserman.
\newblock Very deep convolutional networks for large-scale image recognition.
\newblock {\em arXiv preprint arXiv:1409.1556}, 2014.

\bibitem{tian2016detecting}
Z.~Tian, W.~Huang, T.~He, P.~He, and Y.~Qiao.
\newblock Detecting text in natural image with connectionist text proposal
  network.
\newblock In {\em European Conference on Computer Vision}, pages 56--72.
  Springer, 2016.

\bibitem{veit2016coco}
A.~Veit and e.~a. Matera, Tomas.
\newblock Coco-text: Dataset and benchmark for text detection and recognition
  in natural images.
\newblock {\em arXiv preprint arXiv:1601.07140}, 2016.

\bibitem{wang2010word}
K.~Wang and S.~Belongie.
\newblock Word spotting in the wild.
\newblock In {\em European Conference on Computer Vision}, pages 591--604.
  Springer, 2010.

\bibitem{Yao2012Detecting}
C.~Yao, X.~Bai, W.~Liu, and Y.~Ma.
\newblock Detecting texts of arbitrary orientations in natural images.
\newblock In {\em Computer Vision and Pattern Recognition}, pages 1083--1090,
  2012.

\bibitem{yi2011text}
C.~Yi and Y.~Tian.
\newblock Text string detection from natural scenes by structure-based
  partition and grouping.
\newblock {\em IEEE Transactions on Image Processing}, 20(9):2594--2605, 2011.

\bibitem{Yin2015Multi}
X.~C. Yin, W.~Y. Pei, J.~Zhang, and H.~W. Hao.
\newblock Multi-orientation scene text detection with adaptive clustering.
\newblock {\em IEEE Transactions on Pattern Analysis \& Machine Intelligence},
  37(9):1930, 2015.

\bibitem{zhou2017east}
X.~Zhou, C.~Yao, H.~Wen, Y.~Wang, S.~Zhou, W.~He, and J.~Liang.
\newblock East: An efficient and accurate scene text detector.
\newblock {\em Proceedings of the IEEE Conference on Computer Vision and
  Pattern Recognition}, 2017.

\end{thebibliography}
}

\end{document}